\documentclass[conference]{IEEEtran}
\IEEEoverridecommandlockouts

\usepackage{cite}
\usepackage{amsmath,amssymb,amsfonts}
\usepackage{algorithmic}
\usepackage{algorithm}
\usepackage{graphicx}
\usepackage{textcomp}
\usepackage{xcolor}
\def\BibTeX{{\rm B\kern-.05em{\sc i\kern-.025em b}\kern-.08em
    T\kern-.1667em\lower.7ex\hbox{E}\kern-.125emX}}
\begin{document}

\title{Efficient Identification of High Similarity Clusters in
Polygon Datasets\\

}

\author{\IEEEauthorblockN{John N. Daras}
\IEEEauthorblockA{\textit{Columbia University in the city of New York} \\
New York, USA \\
ind2109@columbia.edu}
}

\maketitle

\begin{abstract}
Advancements in tools like Shapely 2.0 and Triton can
significantly improve the efficiency of spatial similarity computations
by enabling faster and more scalable geometric operations\cite{shapely, triton}. However,
for extremely large datasets, these optimizations may face challenges
due to the sheer volume of computations required. To address this,
we propose a framework that reduces the number of clusters
requiring verification, thereby decreasing the computational load on
these systems. The framework integrates dynamic similarity index
thresholding, supervised scheduling \cite{supervised_giant}, and recall-constrained
optimization to efficiently identify clusters with the highest spatial
similarity while meeting user-defined precision and recall
requirements \cite{recallopt}. By leveraging Kernel Density Estimation (KDE) to
dynamically determine similarity thresholds \cite{parzen} and machine learning
models to prioritize clusters, our approach achieves substantial
reductions in computational cost without sacrificing accuracy.
Experimental results demonstrate the scalability and effectiveness of
the method, offering a practical solution for large-scale geospatial
analysis.
\end{abstract}


\section{Introduction}
Geospatial data constitutes the cornerstone of numerous applications across various domains, including urban planning, environmental monitoring, infrastructure development, and medicine. For example, OpenStreetMap contains global data amounting to over 1.5 terabytes\cite{osm}, while GeoNames describes more than 12 million locations, providing extensive point geometries such as latitude and longitude \cite{geonames}. Expanding these datasets, geospatial knowledge graphs like YAGO2geo integrate millions of lines, polygons, and multipolygons from OpenStreetMap and administrative divisions \cite{yago2geo}, while WorldKG represents around 113.4 million geographic entities \cite{worldkg}. KnowWhereGraph, a more recent initiative, comprises over 12 billion RDF triples, including data on polygons and multipolygons, and supports applications in disaster relief, agricultural land use, and food-related supply chains \cite{knowwheregraph}. Even cross-domain knowledge graphs such as DBpedia and Wikidata incorporate a substantial amount of geospatial information, underscoring the critical role of spatial data on the Web.

Beyond these well-known repositories, spatial datasets also play a transformative role in medicine, particularly in the analysis and modeling of organ structures. For instance, the Visible Human Project provides high-resolution spatial data for anatomical structures \cite{visiblehuman}, while the Human Connectome Project captures detailed spatial relationships within the brain \cite{connectome}. Other datasets, such as the ACDC dataset, focus on cardiac MRI images to segment the heart \cite{acdc}, while the 3D Liver Tumor Dataset provides high-resolution spatial data for liver segmentation and tumor localization \cite{lits}. These datasets are essential for advancing medical imaging, surgical planning, and personalized medicine.

Despite the prominence and growing volume of spatial data, processing and analyzing these datasets efficiently remains a significant challenge. For instance, while OpenStreetMap geometries are widely used, only a fraction of them—0.52\% as of April 2021—are linked to other sources such as Wikidata, highlighting the lack of comprehensive integration across datasets. Similarly, in medical datasets, the complexity and volume of spatial data often lead to computational bottlenecks, particularly when identifying spatial similarities or patterns across large datasets.

Identifying clusters of geometric objects with strong spatial similarities is a fundamental task in geospatial analysis, with applications spanning urban planning, environmental monitoring, infrastructure analysis, medicine, and global datasets. In urban planning, such clusters reveal patterns in land use, infrastructure, or building layouts, aiding targeted development and disaster management. In environmental monitoring, they highlight regions with similar vegetation, soil, or climate characteristics, supporting conservation and agricultural optimization. Medical applications benefit from clustering spatially similar organ shapes or tissue structures to identify abnormalities, plan surgeries, and personalize treatments. Similarly, in datasets like OpenStreetMap, clustering geometrically similar objects improves data integration and interlinking for navigation, disaster response, and supply chain optimization. Across these domains, clustering high-similarity geometries enables meaningful insights, resource optimization, and predictive modeling.

To quantify this, each cluster is assigned a similarity index, calculated as the average pairwise similarity score of its objects. This index serves as a benchmark for determining which clusters exhibit the highest spatial resemblance, providing the foundation for downstream analysis and decision-making. In this work the similarity index is derived from robust spatial similarity metrics, which are designed to capture various geometric and spatial properties of the objects in a cluster. These metrics are grouped into four categories: (1) \emph{Overlap Metrics} (e.g., Jaccard similarity \cite{jaccard}, area similarity); (2) \emph{Shape Complexity Metrics} (e.g., curvature, Fourier descriptors) \cite{polygonmetric, fourier}; (3) \emph{Proportional and Spatial Metrics} (e.g., aspect ratio similarity, bounding box distance) \cite{zhangshape}; and (4) \emph{Boundary Metrics} (e.g., perimeter, circularity \cite{zhangshape}). Together, these metrics provide a comprehensive and nuanced measure of spatial similarity, enabling the accurate calculation of similarity indexes for clusters. However, calculating these similarity metrics for every cluster in a large dataset can be computationally expensive, particularly when the goal is to identify only the top percentage of high-similarity clusters.

Recent advancements in tools like Triton and Shapely~2.0 offer opportunities to improve the efficiency of these calculations. Shapely~2.0 introduces vectorized operations, enabling geometric computations to be performed across multiple shapes simultaneously. By leveraging these capabilities, tasks such as calculating bounding boxes, intersections, and unions for large numbers of clusters can be completed more efficiently, reducing reliance on iterative loops in Python and improving throughput. In addition, Triton provides a way to harness GPU acceleration directly within Python, which can significantly reduce the computational overhead of tasks such as similarity metric calculations once geometric operations have been completed. Despite these optimizations, evaluating similarity metrics across large datasets remains demanding, underscoring the importance of reducing the number of required verifications.

Traditionally, identifying clusters in the top percentage of similarity indexes (e.g., top 10\%, 30\%, or 50\%) would require calculating the similarity indexes for \emph{all} clusters, sorting them, and selecting the desired proportion—an approach that is computationally prohibitive for large datasets. To overcome this limitation, the proposed framework introduces Kernel Density Estimation (KDE) to dynamically determine the similarity index threshold corresponding to the desired top percentage. By estimating the distribution of similarity indexes from a representative sample of clusters, this method identifies the threshold directly, eliminating the need for exhaustive similarity calculations across the entire dataset.

After determining the threshold, the framework employs a supervised scheduling approach to prioritize clusters for verification. During training, a sample of clusters is selected and their similarity indexes are computed. Clusters with similarity indexes above the threshold are labeled as ``high similarity indexed'' (class~1), while those below are labeled as ``low similarity indexed'' (class~0). A machine learning model is then trained using computationally lightweight features paired with these labels, enabling the model to predict which clusters are most likely to exceed the threshold. During verification, the model’s predictions focus computational resources on the most promising candidates, reducing overall cost while maintaining high recall.

To further enhance reliability, a recall-constrained optimization technique is applied to ensure the system achieves a high recall rate. This ensures that the framework reliably identifies all important clusters with high similarity, even if it allows for some false positives. By dynamically adjusting the classification threshold, this technique prioritizes recall over precision so that no high-similarity clusters are overlooked—particularly critical in applications such as urban planning, environmental monitoring, and infrastructure analysis.

\textbf{Contributions.} This study tackles key challenges in identifying top-percentage high-similarity clusters efficiently. Our main contributions are:
\begin{itemize}
  \item Integration of Shapely~2.0 vectorized operations and Triton to enhance the computational efficiency of similarity metric calculations.
  \item A robust methodology for calculating similarity indexes using diverse spatial similarity metrics.
  \item A dynamic thresholding mechanism via KDE for identifying user-specified top-percentage clusters.
  \item A supervised scheduling framework that efficiently classifies and prioritizes clusters.
  \item A recall-constrained optimization approach to achieve high recall while maintaining computational efficiency.
\end{itemize}

Extensive experiments validate the framework’s ability to accurately identify top clusters while significantly reducing computational costs compared to traditional methods. The subsequent sections detail the methodology, supervised scheduling, recall-constrained optimization, and experimental results, highlighting the framework’s scalability and effectiveness.

\section{Related Work}
Identifying high-similarity clusters within large datasets is a challenging task that has drawn insights from multiple research areas, including geospatial analysis, supervised scheduling, and recall-constrained optimization. These fields provide foundational techniques that this work builds upon while addressing unique gaps, such as dynamic similarity index thresholding, machine learning integration, and recall optimization.

The measurement of spatial similarity often forms the cornerstone of cluster analysis. Foundational works rely on metrics such as Jaccard similarity, area overlap, and boundary alignment to quantify spatial resemblance between geometries. Filtering-Verification frameworks have long been a staple in geospatial processing. For instance, Efficient Filtering Algorithms for Location-Aware Publish/Subscribe propose a scalable framework for filtering and verifying spatial objects in location-aware systems. The method divides spatial data into Minimum Bounding Rectangles (MBRs) and uses these rectangles to filter candidate pairs. To further improve performance, the system employs a publish/subscribe mechanism that tracks updates to spatial objects, ensuring that only the relevant pairs are processed. However, while the filtering step reduces computational overhead, the exhaustive verification phase remains computationally expensive for large-scale datasets.

To address scalability challenges, Efficient Privacy-Preserving Spatial Range Query Over Outsourced Encrypted Data introduces a novel method for conducting spatial queries over encrypted datasets. The approach focuses on securely performing range queries while maintaining data confidentiality. The authors propose an efficient spatial indexing mechanism that partitions data into encrypted cells, allowing range-based filtering without decrypting the entire dataset. Although this work primarily addresses privacy concerns, its range query-based filtering emphasizes the importance of efficient indexing for large-scale spatial data, a concept central to high-similarity clustering.

Progressive methods enhance scalability by prioritizing the processing of candidate pairs, particularly in resource-constrained environments. Supervised Progressive GIA.nt extends the Filtering-Verification paradigm by introducing a machine learning-based scheduling component. The system trains a lightweight model using features such as area overlap, boundary alignment, and aspect ratios to assign scores to candidate pairs. These scores determine the verification order, ensuring that pairs with higher probabilities of topological relationships are processed first. By leveraging a supervised learning approach, this method reduces runtime while maintaining high recall. However, its focus on pairwise interlinking tasks rather than clustering limits its applicability to the challenges addressed in this paper.

Parallel processing frameworks also play a critical role in handling massive spatial datasets. Big Spatial Data Processing Frameworks: Feature and Performance Evaluation evaluates the performance of distributed systems, such as Apache Spark, in processing spatial queries. The study compares various data partitioning strategies, including spatial R-trees and grid-based indexing, to optimize query execution time. The results highlight the importance of selecting appropriate indexing mechanisms for different dataset characteristics. While this work underscores the utility of parallelism, its focus on spatial queries rather than similarity-based clustering leaves room for improvement in cluster-level analyses.

Dynamic thresholding techniques have been applied in domains requiring adaptive decision-making under data variability. For instance, Dynamic Thresholding and Recall-Constrained Optimization for Environmental Monitoring employs Kernel Density Estimation (KDE) to determine dynamic thresholds for identifying critical environmental events. By analyzing a sample of data points, the system estimates the distribution of key features, allowing thresholds to be adjusted dynamically. This reduces computational overhead while ensuring that significant events are not overlooked. The KDE-based dynamic thresholding mechanism aligns closely with this study's methodology for determining similarity index thresholds.

Recall-constrained optimization is essential in applications where missing critical regions or clusters can have severe consequences. Ecological Thresholds: The Key to Successful Environmental Management or an Important Concept with No Practical Application explores the trade-offs between precision and recall in ecological monitoring. The study demonstrates that fixed thresholds often fail to capture dataset variability, leading to missed ecological changes. By incorporating adaptive thresholding, this work highlights the need for recall-focused strategies in data-driven decision-making. Similarly, Iterated Dynamic Thresholding Search for Packing Equal Circles into a Square introduces iterative threshold adjustments to optimize spatial layouts, ensuring that critical relationships between objects are preserved. This iterative mechanism provides insights into balancing computational efficiency with spatial accuracy.

Recent advancements in computational tools, such as Shapely~2.0 and Triton, present opportunities to improve the efficiency of spatial similarity calculations in large-scale datasets. Shapely~2.0 introduces vectorized geometric operations, enabling the parallel processing of tasks like intersection, union, and area calculations. Triton facilitates high-performance GPU-accelerated computations for numerically intensive tasks, such as feature aggregation and normalization, making it especially valuable for handling large datasets. Prior studies, such as those on GPU-accelerated geometry and scalable spatial data mining, have demonstrated the potential of these technologies to enhance performance. However, the computational burden remains significant, underscoring the importance of reducing the number of required verifications. By combining these tools with dynamic thresholding and supervised scheduling, this work ensures computational efficiency while maintaining scalability and precision.

While prior work in clustering, supervised scheduling, and recall optimization has individually advanced these fields, no existing framework integrates these methods for the specific purpose of identifying high-similarity clusters efficiently. This paper addresses this gap by combining dynamic similarity index thresholding using KDE, supervised scheduling with machine learning, and recall-constrained optimization into an end-to-end solution. Unlike existing works, this approach dynamically determines a similarity index threshold tailored to user-specified top percentages, trains a machine learning model on computationally efficient features to prioritize clusters, and applies recall-constrained optimization to ensure all critical clusters are identified. This integrated methodology fills a key gap in both geospatial and clustering literature, offering a scalable and robust solution for applications in geospatial analysis, environmental monitoring, and medical imaging.

\section{Preliminaries}
\label{sec:prelim}

\newcommand{\area}[1]{\lvert #1\rvert}           
\newcommand{\perim}[1]{P(#1)}                    
\newcommand{\cent}[1]{c(#1)}                     
\newcommand{\fd}[1]{\mathbf{F}(#1)}              
\newcommand{\aratio}[1]{r(#1)}                   
\newcommand{\circul}[1]{\phi(#1)}                

\subsection{Mathematical Formulation of the Similarity Index}
The  similarity  index  quantifies  the  spatial  resemblance
ofobjects within a cluster. It is defined as the average pairwise
similarity score between all objects in a cluster C:  
\begin{equation}
\mathrm{SI}(C)= \frac{2}{|C|(|C|-1)} \sum_{1\le i<j\le |C|} \mathrm{Similarity}(O_i,O_j).
\end{equation}
where:  
\begin{itemize}
  \item $|C|$ is the number of objects in the cluster.
  \item $O_i$ and $O_j$ are geometric objects within the cluster.
  \item Similarity$(O_i, O_j)$ is the pairwise similarity score, computed
using the spatial  similarity metrics  described  in  Section 4.2.
This  index  provides  a  single  value  summarizing  the  spatial
consistency of a cluster. Clusters with higher similarity indexes
exhibit  stronger  spatial  alignment,  uniformity  in  shape,  and
proportional characteristics.
\end{itemize}

\subsection{Spatial Similarity Metrics}

\paragraph{Polygon Centering}
Before  computing  pairwise  similarities,  all  polygons  are
preprocessed  to  ensure  that  their  positions  in  space  do  not
influence similarity calculations. This involves translating each
polygon such that its centroid aligns with the origin. The steps
include:

1. Centroid Calculation: For each polygon, its centroid is
determined.

2.  Translation:  The polygon is  translated  by shift  ing its
centroid to the origin using affine transformations.

Purpose:  This  preprocessing  ensures  that  spatial  translations
(e.g., the location of polygons in a global coordinate system) do
not  affect  similarity  computations.  By  focusing  solely  on
intrinsic geometric and spatial  characteristics,  such as shape,
boundary complexity, and area,  this step ensures consistency
and fairness in similarity calculations.

Rationale: The absolute position of a polygon is irrelevant for
many  applications,  such  as  clustering  or  pattern  analysis.
Centering  allows  the  framework  to  prioritize  geometric
properties,  making it particularly effective for applications in
urban  planning,  environmental  monitoring,  and  medical
imaging

\begin{figure}[h]
\centering
\includegraphics[width=0.73\linewidth,height=0.11\textheight,keepaspectratio=false]{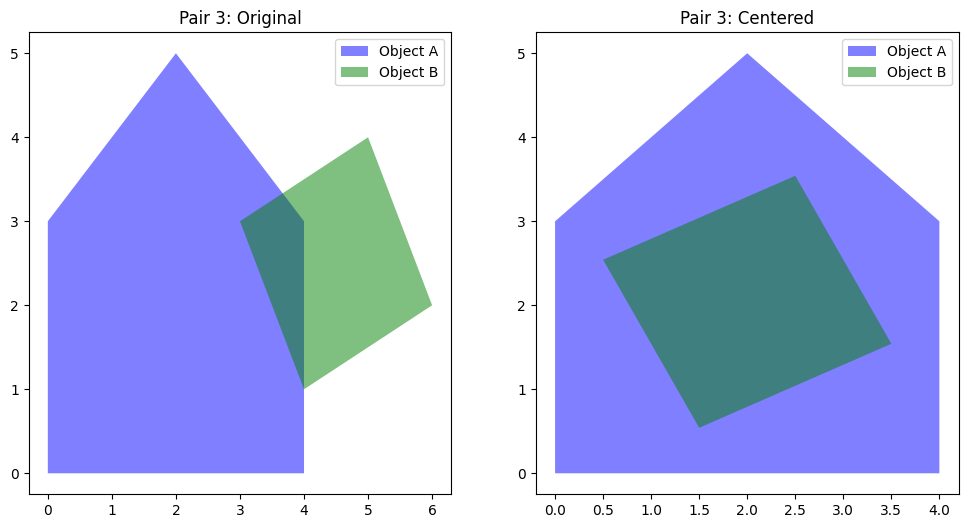}
\caption{\small Centering polygons ensures spatial positioning
does not influence similarity metrics.}
\end{figure}

\medskip
The pairwise similarity score Similarity$(O_i, O_j)$ is computed
using a combination of metrics that capture different geometric
and spatial properties of polygons. These metrics include:  :contentReference[oaicite:2]{index=2}

\begin{enumerate}
  \item \textbf{Jaccard Similarity}\\
  \textit{Purpose:} Measures the overlap between two polygons.\\
  \textit{Why  Used:}  Highlights  overlap-based  relationships  while
penalizing disjoint or minimally overlapping shapes, making it
critical for identifying clusters with shared spatial regions.\\
  \textit{Formula:}
  \begin{equation}
  J(A,B)=\frac{\area{A\cap B}}{\area{A\cup B}}.
  \end{equation}

  \item \textbf{Area Similarity}\\
  \textit{Purpose:}  Emphasizes  proportional  overlap  by accounting for
the relative scales of the polygons.\\
  \textit{Why  Used:}  Balances  similarity  evaluation  for  polygons  of
varying  sizes,  unlike  Jaccard  similarity,  which  can  under
emphasize smaller overlaps.\\
  \textit{Formula:}
  \begin{equation}
  S_{\text{area}}(A,B)=\frac{2\,\area{A\cap B}}{\area{A}+\area{B}}.
  \end{equation}

  \item \textbf{Curvature Similarity}\\
  \textit{Purpose:}  Quantifies  boundary  complexity,  differentiating
between simple and intricate shapes.\\
  \textit{Why Used:} Ensures that clusters are sensitive to the level of
detail  in  object  boundaries,  which  is  essential  for  complex
geometric objects.\\
  \textit{Formula:}
  \begin{equation}
  S_{\text{curv}}(A,B)=\exp\!\left(-\frac{\lvert n_A-n_B\rvert}{\max(n_A,n_B)}\right).
  \end{equation}

  \item \textbf{Fourier Descriptor Similarity}\\
  \textit{Purpose:}  Encodes  global  shape  geometry  into  a  frequency
domain.\\
  \textit{Why  Used:}  Provides  robustness  against  minor  distortions,
making it ideal for comparing shapes with slight variations.\\
  \textit{Formula:}
  \begin{equation}
  S_{\text{FD}}(A,B)=\frac{1}{1+\lVert \fd{A}-\fd{B}\rVert}.
  \end{equation}

  \item \textbf{Aspect Ratio Similarity}\\
  \textit{Purpose:}  Reflects  differences  in  elongation or  orientation  of
bounding boxes.\\
  \textit{Why Used:}  Adds  a  proportionality-based  perspective,  useful
for distinguishing between elongated and compact shapes.\\
  \textit{Formula:}
  \begin{equation}
  S_{\text{ar}}(A,B)=\frac{1}{1+\lvert \aratio{A}-\aratio{B}\rvert}.
  \end{equation}

  \item \textbf{Perimeter Similarity}\\
  \textit{Purpose:} Focuses on the boundary lengths of polygons.\\
  \textit{Why Used:} Captures size-related differences and complements
area-based metrics.\\
  \textit{Formula:}
  \begin{equation}
  S_{\text{perim}}(A,B)=\frac{1}{1+\lvert \perim{A}-\perim{B}\rvert}.
  \end{equation}

  \item \textbf{Bounding Box Distance}\\
  \textit{Purpose:} Measures the spatial proximity between bounding box
centers.\\
  \textit{Why Used:}  Crucial  for  clustering applications where  nearby
objects are more likely to be related.\\
  \textit{Formula:}
  \begin{equation}
  S_{\text{bb}}(A,B)=\frac{1}{1+\lVert \cent{A}-\cent{B}\rVert_2}.
  \end{equation}

  \item \textbf{Polygon Circularity Similarity}\\
  \textit{Purpose:}  Highlights  differences  in  roundness  between
polygons.\\
  \textit{Why Used:} Distinguishes between regular shapes (e.g., circles)
and  irregular  ones,  providing  nuanced  insights  into  shape
characteristics.\\
  \textit{Formulas:}
  \begin{equation}
  \circul{A}=\frac{4\pi\,\area{A}}{\perim{A}^{\,2}},\qquad
  S_{\text{circ}}(A,B)=\frac{1}{1+\lvert \circul{A}-\circul{B}\rvert}.
  \end{equation}

  \item \textbf{Combined Similarity}\\
  \textit{Purpose:} Aggregates the above metrics into a single similarity
score using weighted contributions.\\
  \textit{Why Used:} Balances individual metrics to deliver a robust and
comprehensive similarity measure.\\
  \textit{Formula:}
  \begin{equation}
  s(A,B)=\sum_i w_i\,m_i(A,B).
  \end{equation}
\end{enumerate}

\subsection{Supervised Scheduling}
Supervised  scheduling  is  a  key  optimization  technique  em-
ployed in this  framework  to  identify  high-similarity  clusters
efficiently.  Rather  than  exhaustively  calculating  similarity
indexes  for  all  clusters,  supervised  scheduling  leverages
machine  learning  to  prioritize  clusters  likely  to  exceed  the
similarity index threshold.

In this framework, a machine learning model is trained using
computationally  lightweight  features  extracted  from  clusters
and  binary  labels  indicating  whether  a  cluster  exceeds  the
similarity index threshold. During prediction, the model assigns
a likelihood score to each cluster, allowing the system to focus
on the most promising candidates.

This approach significantly reduces computational costs while
ensuring high recall, making it particularly effective for large
datasets  and  applications  where  computational  efficiency  is
paramount. A detailed description of the supervised scheduling
algorithm is provided in Section 4.  

\subsection{Kernel Density Estimation for Similarity Index 
Thresholding}
Kernel  Density  Estimation  (KDE)  is  a  non-parametric
statistical  method  used  to  estimate  the  probability  density
function (PDF) of a dataset. In this framework, KDE plays a
pivotal  role  in  dynamically  determining  the  similarity  index
threshold, which defines the boundary between high-similarity
and low-similarity clusters.

Purpose:  The  similarity  index  threshold  is  essential  for
identifying  clusters  within  the  top  percentage  of  similarity
indexes, as specified by the user. For example, to identify the
top 10\% of clusters, the threshold is set such that only 10\% of
similarity indexes in the dataset are above it.

How KDE is Applied: A representative sample of clusters is
selected from the dataset. Similarity indexes are calculated for
this  sample  using  robust  spatial  similarity  metrics.  KDE  is
applied  to  estimate  the  probability  density  function  of  the
similarity indexes, providing a smooth approximation of their
distribution. Finally, the threshold is determined by finding the
similarity index value such that the area under the PDF above
this value corresponds to the desired percentage (e.g., 10\% or
30\%).

Advantages  of  KDE:  KDE offers  several  advantages  in  this
framework.  First,  it  dynamically  adapts  to  the  underlying
distribution of similarity indexes,  allowing the framework to
handle  diverse  datasets  without  relying  on  static  thresholds.
Second, it is computationally efficient because it estimates the
threshold from a representative sample, eliminating the need to
compute similarity indexes for the entire dataset. Finally, KDE
provides  flexibility,  supporting  varying  user-defined
percentages, making it applicable across different use cases.

KDE  ensures  that  the  threshold  dynamically  adapts  to  the
datasetˆas  characteristics  while  maintaining  computational
efficiency. A detailed explanation of the KDE process and its
integration into the overall framework is provided in Section 4.  

\newcommand{\algshrink}{\fontsize{8}{9.2}\selectfont}  
\newcommand{\lnum}[1]{\textbf{\scriptsize #1:}}        
\setlength{\textfloatsep}{8pt plus 2pt minus 2pt}
\setlength{\floatsep}{6pt plus 2pt minus 2pt}
\catcode`\^^I=10\relax

\section{APPROACH}
\label{sec:approach}

\subsection{Features for Supervised Scheduling}

For each cluster, we define a representative geometry $g$, which
serves as a reference for computing cluster-wide properties.
Geometry $g$ is either a polygon or a linestring, chosen such that
its envelope (bounding box) intersects with the envelopes of all
other geometries in the cluster. This ensures that geometry $g$
encapsulates the spatial footprint of the cluster, providing a
unified basis for feature computation.

The representative geometry $g$ plays a central role in
simplifying the characterization of clusters. By relying on a
single geometry that spatially relates to all cluster members, the
framework can compute cluster-wide features efficiently while
maintaining consistency across diverse datasets.

\begin{figure}[h]
\centering
\includegraphics
[width=0.73\linewidth,height=0.11\textheight,keepaspectratio=false]{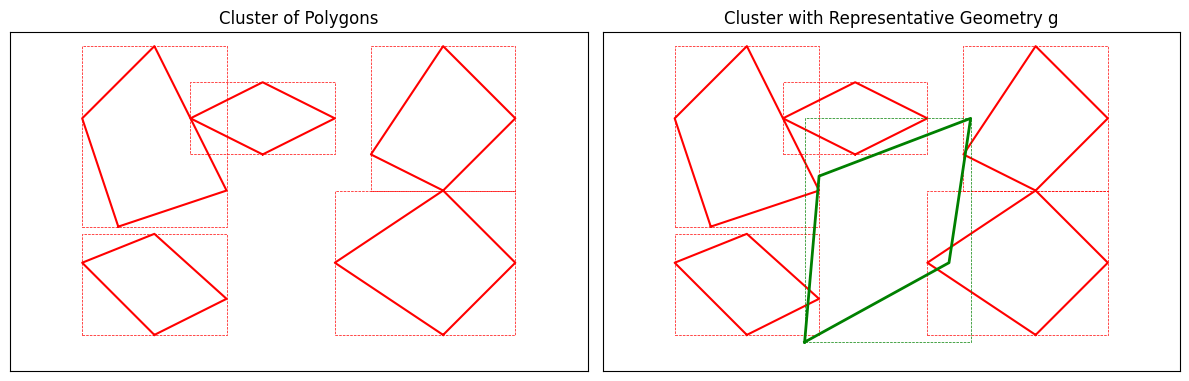}
\caption{Cluster’s representative geometry $g$.}
\end{figure}

The supervised scheduling framework relies on a
comprehensive set of features to represent each cluster. These
features are extracted from:

\begin{enumerate}
    \item \textbf{Source Geometries.} These are the geometries within the
    cluster.
    \item \textbf{Target Geometry.} This is the representative geometry $g$,
    defined as the geometry whose envelope intersects with the
    envelopes of all geometries in the cluster.
\end{enumerate}

The extracted features are lightweight, focusing on
computational efficiency while still capturing critical spatial
and geometric properties of the clusters. Lightweight features
are essential to reduce computational overhead, enabling the
machine learning model to operate effectively on large datasets
without compromising accuracy or scalability. The 15 features
we extract are inspired from the paper \textit{Supervised Progressive
GIA.nt}~\cite{supervised_giant}. We extract the following features for
each pair of source geometry (cluster geometry) and target
geometry (representative geometry $g$):

\begin{itemize}
    \item[F1)] Source MBR (Minimum Bounding Rectangle) area.
    \item[F2)] Target MBR area (envelope area of the representative geometry $g$).
    \item[F3)] Number of grid blocks occupied by the source geometry’s MBR.
    \item[F4)] Number of grid blocks occupied by the target geometry’s MBR.
    \item[F5)] Frequency of overlaps (how often the source geometry overlaps with other geometries).
    \item[F6)] Number of points (vertices) in the source geometry.
    \item[F7)] Number of points (vertices) in the target geometry.
    \item[F8)] Perimeter (total boundary length) of the source geometry.
    \item[F9)] Perimeter (total boundary length) of the target geometry.
    \item[F10)] Total co-occurrences (number of geometries that overlap with the source geometry’s envelope).
    \item[F11)] Distinct co-occurrences (unique geometries overlapping with the source geometry’s envelope).
    \item[F12)] Real candidate count (valid overlapping geometries that meet a predefined spatial relationship).
    \item[F13)] Sum of overlap frequencies for candidate geometries.
    \item[F14)] Count of candidate geometries overlapping with the target geometry.
    \item[F15)] Number of geometries within the cluster (cluster size).
\end{itemize}

\subsubsection*{Normalization Process}

Normalization ensures that features extracted from different
clusters and geometries are scaled consistently, enabling fair
comparisons and efficient learning. Two distinct sets of
minimum and maximum values are used in this framework.

\textbf{Min-Max Values for Individual Geometries} are the minimum
and maximum values of each feature $F1$ to $F15$, computed
across all individual geometries in the dataset. They define the
range for normalizing individual features for each geometry.
For example, the minimum value for the envelope area (F1)
corresponds to the smallest envelope area across all geometries
in the dataset, while the maximum value corresponds to the
largest. The normalization formula for individual geometries is:

\[
NF_{i,j} = \frac{FV_{i,j} - \min Val_j}{\max Val_j} \cdot 10000
\]

where:
\begin{itemize}
    \item $NF_{i,j}$ is the normalized feature for geometry $i$ and feature $j$.
    \item $FV_{i,j}$ is the feature value for geometry $i$ and feature $j$.
    \item $\min Val_j$ is the minimum value of feature $j$ across all geometries.
    \item $\max Val_j$ is the maximum value of feature $j$ across all geometries.
\end{itemize}

\textbf{Min-Max Values for Clusters (Cluster-Level Min-Max)} are the
minimum and maximum values of aggregated cluster features,
computed across all clusters in the dataset. Aggregated features
(mean values for each feature in a cluster) are normalized using
these values to ensure comparability across clusters. The
formula for cluster-level normalization is:

\[
NCF_j = \frac{\overline{NF_j} - \min Cf_j}{\max Cf_j} \cdot 10000
\]

where:
\begin{itemize}
    \item $NCF_j$ is the normalized cluster feature for feature $j$.
    \item $NF_j$ is the mean of normalized feature $NF_{i,j}$ across all
    geometries in the cluster.
    \item $\min Cf_j$ is the minimum value of the aggregated feature $j$ across all clusters.
    \item $\max Cf_j$ is the maximum value of the aggregated feature $j$ across all clusters.
\end{itemize}

After applying the above process, we derive a total of 16 final
features for each cluster. These normalized features will later
be used to train the machine learning model.

Purpose of Normalization: The two-step formulas are designed to improve the accuracy of the machine learning model by ensuring that features are
consistent across scales. This helps to reduce the impact of
variations in raw feature magnitudes, which can arise from the
diverse sizes, shapes, and scales of geometries in the dataset. By
applying these formulas, the features can be compared in a
more uniform way. At the individual geometry level, the
formulas help features capture meaningful patterns while
reducing the influence of extreme values. At the cluster level,
the formulas ensure that aggregated features incorporate
information not only from within the cluster but also relative to
the overall range of values observed across all clusters in the
dataset. This broader perspective helps the model identify
clusters with high similarity indices more effectively.

\subsection{Algorithm}

Cluster Finding (lines 1--24):
The Cluster Finding step organizes geometries into clusters and
prepares data for subsequent phases of the algorithm. This step
involves two main tasks: indexing the source geometries and
identifying candidate clusters by associating target geometries
with relevant source geometries. 

First, an index $I$ is initialized to store the spatial positions of
source geometries. The granularity of the index grid is
determined by the function $DefineIndexGranularity(S)$,
which computes the grid resolution based on the properties of
the source dataset $S$. Each geometry $s \in S$ is added to the
index $I$ using $AddToIndex(I,s)$. This indexing enables efficient
spatial queries in subsequent steps. 

Next, several data structures are initialized. The total number
of potential geometry pairs is computed as $D = |S| \times |T|$,
where $|S|$ and $|T|$ are the sizes of the source and target
datasets, respectively. The $sourceStats$ structure stores
statistics for each source geometry, such as the number of
distinct and real pairs. Two subsets, $sample$ and
$verification\_sample$, are created using the function
$RandomGenerator(m, D)$ to randomly select geometry pairs
for KDE modeling and validation.

\setlength{\textfloatsep}{20pt plus 2pt minus 2pt} 
\setlength{\intextsep}{15pt plus 2pt minus 2pt}    

\vspace{\baselineskip}
\vspace{\baselineskip}

\begin{algorithm}
\caption{Algorithm}
\label{alg:hsched-part1}

\newlength{\algheaderindentI}
\settowidth{\algheaderindentI}{\lnum{1}\ } 
{\algshrink
\begingroup
\setlength{\leftskip}{\algheaderindentI}
\noindent 
\textbf{Input}: the source dataset $S$, the target dataset $T$, the feature set $F$, the\\
maximum sample size $m$, the class size $N$, the probabilistic\\
classification algorithm $A$, desired recall $r_d$, similarity range $\textit{similarity\_range}$,\\
\textbf{Output}: the links $L_R = \{ (c, s ) \subseteq C \times S : s(c) \}$\par
\endgroup
}


\begin{algorithmic}
\algshrink
\STATE \lnum{1} $I \gets \varnothing$;\quad $(\Delta x,\Delta y)\gets \mathrm{DefineIndexGranularity}(S)$
\STATE \lnum{2} \textbf{foreach} geometry $s\in S$ \textbf{do} \hfill \texttt{// filtering}
\STATE \lnum{3} \quad $\mathrm{AddToIndex}(I,s)$
\STATE \lnum{4} \textbf{end}
\STATE \lnum{5} $D\gets |S|\cdot|T|$;\ \textit{sourceStats}$\gets\varnothing$;\ $id\gets 0$;\ \textit{sample}$\gets\varnothing$;\ \textit{verification\_sample}$\gets\varnothing$
\STATE \lnum{6} \textit{sampleIds}$\gets \mathrm{RandomGenerator}(m,D)$
\STATE \lnum{7} \textbf{foreach} geometry $t\in T$ \textbf{do} \hfill \texttt{// first pass}
\STATE \lnum{8} \quad $CS\gets\varnothing$ \hfill \texttt{// set of source candidates}
\STATE \lnum{9} \quad $(x_1(t),y_1(t),x_2(t),y_2(t))\gets \mathrm{GetDiagCorners}(t)$
\STATE \lnum{10} \quad \textbf{for} $i\gets \lfloor x_1(t)\Delta x\rfloor$ \textbf{to} $\lceil x_2(t)\Delta x\rceil$ \textbf{do}
\STATE \lnum{11} \qquad \textbf{for} $j\gets \lfloor y_1(t)\Delta y\rfloor$ \textbf{to} $\lceil y_2(t)\Delta y\rceil$ \textbf{do}
\STATE \lnum{12} \qquad\quad $CS\gets CS\cup \mathrm{GetTileContents}(I,i,j)$
\STATE \lnum{13} \qquad \textbf{end}
\STATE \lnum{14} \quad \textbf{end}
\STATE \lnum{15} \quad \textbf{foreach} geometry $s\in CS$ \textbf{do}
\STATE \lnum{16} \qquad \textit{sourceStats}$\gets \mathrm{UpdateTotalDistinctPairs}(\textit{sourceStats},s)$
\STATE \lnum{17} \qquad \textbf{if} $\mathrm{IntersectingMBRs}(s,t)$ \textbf{then}
\STATE \lnum{18} \qquad\quad \textit{sourceStats}$\gets \mathrm{UpdateRealPairs}(\textit{sourceStats},s)$;\quad \textit{Candidates}$\gets \textit{Candidates}\cup\{s\}$
\STATE \lnum{19} \qquad \textbf{end}
\STATE \lnum{20} \qquad \textbf{if} $id\in \textit{sampleIds}$ \textbf{then} \textit{sample}$\gets \textit{sample}\cup\{(s,t)\}$ \textbf{;}\  \textit{kde\_sample}$\gets \textit{kde\_sample}\cup\{(s,t)\}$ \ \ \texttt{// do the same for kde\_sample}
\STATE \lnum{21} \qquad $id\gets id+1$
\STATE \lnum{22} \quad \textbf{end}
\STATE \lnum{23} \quad $\mathrm{AddCluster}(\textit{allClusters},\textit{Candidates})$
\STATE \lnum{24} \textbf{end}
\end{algorithmic}
\end{algorithm}

\vspace{\baselineskip}
\vspace{\baselineskip}

The algorithm then iterates over each target geometry $t \in T$
to identify candidate clusters. For each target geometry $t$, a
set $CS$ is initialized to store potential source geometry
candidates whose bounding boxes intersect with the bounding
box of $t$. The diagonal corners of $t$’s bounding box,
$(x_1, y_1)$ and $(x_2, y_2)$, are computed using
$GetDiagCorners(t)$. The bounding box is then divided into
grid cells, and for each grid cell, the contents of the
corresponding tiles in the index $I$ are added to $CS$. 

For each geometry $s \in CS$, the $sourceStats$ structure is
updated to track the total number of distinct pairs involving $s$
using $UpdateTotalDistinctPairs(sourceStats, s)$. If the bounding
boxes of $s$ and $t$ intersect, the $sourceStats$ structure is
further updated with the count of real pairs using
$UpdateRealPairs(sourceStats, s)$, and the geometry $s$ is
added to a set of candidates for the cluster. If the pair ID is
part of the randomly generated sample set, the pair $(s,t)$ is
added to both the $sample$ and $kde\_sample$ subsets for
subsequent analysis. Finally, the set of candidate geometries
for the current target geometry $t$ is added to $allClusters$,
which stores all identified clusters. This prepares the dataset
for further processing, including the KDE modeling and training
phases. 

The Cluster Finding step is crucial for efficiently associating
source and target geometries. By leveraging spatial indexing
and grid-based filtering, the algorithm focuses only on relevant
candidate geometries, reducing computational overhead and
facilitating the identification of meaningful clusters for later
stages of the process.

KDE Phase to Find Minimum Similarity Threshold (lines 25--29):
The KDE phase estimates the minimum similarity threshold
required to identify clusters with high similarity indices. This
step begins by iterating through all clusters in the
$kde\_sample$, which is a representative subset of the dataset.
For each cluster, the similarity index is computed using the
function $SimilarityCalculator(Cluster)$. The computed
similarity indices are stored in a list.

Once all similarity indices are collected, a Kernel Density
Estimation (KDE) model is applied to fit the distribution of
similarity indices using the $GetBestModel(List)$ function. KDE
provides a smooth estimation of the underlying probability
density function, enabling precise identification of thresholds.
Using this model, the $EstimateThreshold(KDE,
similarity\_range)$ function determines the minimum similarity
threshold, which corresponds to the desired similarity range
specified by the user (e.g., top 10\% or top 50\% similarity
indices). This threshold will be used as a benchmark in the
subsequent training and verification phases.

Training Phase (lines 30--38): The training phase aims to label clusters and train a machine
learning model to classify clusters efficiently. Two sets,
$negClusters$ (negative clusters) and $posClusters$ (positive
clusters), are initialized to store clusters with similarity indices
below and above the threshold, respectively. The $sample$ and
$verification\_sample$ are shuffled to ensure a balanced and
unbiased representation of the dataset. 

The algorithm iterates through the clusters in the $sample$. For
each cluster, the similarity index is computed using
$SimilarityCalculator(Cluster)$. If the similarity index is greater
than or equal to the minimum similarity threshold, the cluster is
added to $posClusters$; otherwise, it is added to $negClusters$.
This process continues until both sets contain at least $N$
clusters, ensuring a sufficient number of labeled samples for
training. 

Once the clusters are labeled, features are extracted from the
union of $posClusters$ and $negClusters$ using the
$GetFeatures(posClusters \cup negClusters, F, sourceStats, I)$
function. This step generates a comprehensive feature set for
training the machine learning model. The labeled data is then
used to train the probabilistic classifier $M$ using the
$Train(L)$ function, where $L$ represents the labeled dataset
with extracted features. After training, key data structures are
initialized for the subsequent verification phase: $L_R$, which
will store the final results; $minw$, which tracks the minimum
weight; $TC$, which stores total clusters; and $PQ$, a priority
queue that ranks clusters based on their predicted classification
probabilities.

\vspace{\baselineskip}
\vspace{\baselineskip}
\vspace{\baselineskip}

\begin{algorithm}

\label{alg:hsched-part2}
\begin{algorithmic}
\algshrink
\STATE  \texttt{// KDE phase to find minimum similarity threshold}
\STATE \lnum{25} \textbf{foreach} \textit{Cluster} $\in \textit{kde\_sample}$ \textbf{do}
\STATE \lnum{26} \quad $\ell \gets \mathrm{SimilarityCalculator}(\textit{Cluster})$
\STATE \lnum{27} \quad $\mathrm{ListAdd}(\textit{List},\ell)$
\STATE \lnum{28} $\mathrm{KDE} \gets \mathrm{Get\_Best\_Model}(\textit{List})$
\STATE \lnum{29} $\textit{min\_sim\_thr} \gets \mathrm{Estimate\_Threshold}(\mathrm{KDE},\,\mathrm{similarity\_range})$

\STATE \lnum{30} $\textit{negClusters}\gets\varnothing$;\  $\textit{posClusters}\gets\varnothing$;\  $\mathrm{Shuffle}(\textit{sample})$;\  $\mathrm{Shuffle}(\textit{verification\_sample})$
\STATE \lnum{31} \textbf{foreach} \textit{Cluster} $\in \textit{sample}$ \textbf{do} \hfill \texttt{// labelling}
\STATE \lnum{32} \quad $\ell \gets \mathrm{SimilarityCalculator}(\textit{Cluster})$
\STATE \lnum{33} \quad \textbf{if} $\ell \ge \textit{min\_sim\_thr}$ \textbf{then} $\textit{posClusters}\gets \textit{posClusters}\cup\{\textit{Cluster}\}$
\STATE \lnum{34} \quad \textbf{else} $\textit{negClusters}\gets \textit{negClusters}\cup\{\textit{Cluster}\}$
\STATE \lnum{35} \quad \textbf{if} $N \le |\textit{posClusters}|$ \textbf{and} $N \le |\textit{negClusters}|$ \textbf{then break}
\STATE \lnum{36} \textbf{end}
\STATE \lnum{37} $L \gets \mathrm{GetFeatures}\big(\textit{posClusters}\cup\textit{negClusters},\,F,\,\textit{sourceStats},\,I\big)$
\STATE \lnum{38} $M \gets \mathrm{Train}(L)$;\  $L_R\gets\varnothing$;\  $\textit{minw}\gets 0$;\  $\textit{TC}\gets\varnothing$;\  $\textit{PQ}\gets\varnothing$ \ \ \texttt{// PQ is the priority queue}

\STATE \lnum{39} \textbf{foreach} \textit{Cluster} $\in \textit{kde\_sample}$ \textbf{do}
\STATE \lnum{40} \quad $u \gets \mathrm{GetFeatureVector}(\textit{Cluster},F)$
\STATE \lnum{41} \quad $w_{s,t} \gets \mathrm{GetClassificationProbability}(M,\,u)$
\STATE \lnum{42} \quad $\mathrm{PQAdd}(\textit{PQ},\,\textit{Cluster},\,w_{s,t})$
\STATE \lnum{43} \textbf{end}
\STATE \lnum{44} $\textit{PQ.size} \gets N$
\STATE \lnum{45} $\textit{HighSimIndices} \gets \mathrm{NumberAboveSimilarityThreshold}(\textit{PQ})$
\STATE \lnum{46} \textbf{while} $\textit{PQ} \neq \varnothing$ \textbf{do}
\STATE \lnum{47} \quad $c \gets \mathrm{PQPopLast}(\textit{PQ})$ \hfill \texttt{// remove top-weighted cluster}
\STATE \lnum{48} \quad \textbf{if} $\mathrm{SimilarityCalculator}(c) \ge \textit{min\_sim\_thr}$ \textbf{then}
\STATE \lnum{49} \qquad $\textit{highSimilarityPrediction} \gets \textit{highSimilarityPrediction} + 1$
\STATE \lnum{50} \quad \textbf{end}
\STATE \lnum{51} \quad \textbf{if} $\textit{counter} = \mathrm{similarity\_range}\cdot N$ \textbf{then break}
\STATE \lnum{52} \quad $\textit{counter} \gets \textit{counter} + 1$
\STATE \lnum{53} \textbf{end}

\STATE \lnum{57} $\textit{recall\_approx} \gets \textit{highSimilarityPrediction} \,/\, \textit{HighSimIndices}$
\STATE \lnum{58} $\textit{max\_size} \gets r_d \cdot (1/\textit{recall\_approx}) \cdot (\textit{HighSimIndices}/N) \cdot \textit{totalClusters}$

\STATE \lnum{59} $\textit{PQ}.\mathrm{empty}$ \hfill \texttt{// reset queue}
\STATE \lnum{60} \textbf{foreach} \textit{Cluster} $\in \textit{allClusters}$ \textbf{do}
\STATE \lnum{61} \quad $u \gets \mathrm{GetFeatureVector}(\textit{Cluster},F)$
\STATE \lnum{62} \quad $w_{s,t} \gets \mathrm{GetClassificationProbability}(M,\,u)$
\STATE \lnum{63} \quad $\mathrm{PQAdd}(\textit{PQ},\,\textit{Cluster},\,w_{s,t})$
\STATE \lnum{64} \textbf{end}
\STATE \lnum{65} \textbf{while} $\textit{PQ} \neq \varnothing$ \textbf{do}
\STATE \lnum{66} \quad $c \gets \mathrm{PQPopLast}(\textit{PQ})$
\STATE \lnum{67} \quad $I_M \gets \mathrm{SimilarityCalculator}(c)$
\STATE \lnum{68} \quad \textbf{if} $I_M \ge \textit{min\_sim\_thr}$ \textbf{then}
\STATE \lnum{69} \qquad $L_R \gets L_R \cup \{c\}$
\STATE \lnum{70} \qquad $\textit{highSimilarityPrediction} \gets \textit{highSimilarityPrediction} + 1$
\STATE \lnum{71} \quad \textbf{end}
\STATE \lnum{72} \quad \textbf{if} $\textit{counter} = \textit{max\_size}$ \textbf{then break}
\STATE \lnum{73} \quad $\textit{counter} \gets \textit{counter} + 1$
\STATE \lnum{74} \textbf{end}

\STATE \lnum{78} \textbf{return} $L_R$
\end{algorithmic}
\end{algorithm}

\vspace{\baselineskip}
\vspace{\baselineskip}

The KDE and training phases are critical to efficiently
identifying high-similarity clusters. KDE reduces computational
overhead by estimating a data-driven threshold, while the
training phase equips the framework with a model capable of
classifying clusters based on lightweight, extracted features.

Estimate Recall and Simulate Verification Process in a Small Sample (lines 39--58):

This phase evaluates the system’s ability to identify
high-similarity clusters and estimates the recall under simulated
verification. The process begins by iterating through all clusters
in the $kde\_sample$, a representative subset of the dataset. For
each cluster, the feature vector $u$ is generated using the
$GetFeatureVector(Cluster, F)$ function, which extracts the
lightweight features defined earlier. The trained classifier $M$ is
then used to compute the classification probability $w_{s,t}$ for the
cluster. This probability represents the likelihood that the
cluster belongs to the high-similarity category. Each cluster,
along with its classification probability, is added to a priority
queue $PQ$, which ranks clusters by their weights $w_{s,t}$. The
size of $PQ$ is limited to $N$, the desired number of top clusters
to process. The number of clusters with similarity indices above
the minimum similarity threshold is calculated using the
$NumberAboveSimilarityThreshold(PQ)$ function, which
identifies the count of clusters classified as highly similar.

Next, the algorithm simulates the verification process using the
priority queue. Clusters are processed in descending order of
their classification probabilities by repeatedly removing the
highest-ranked cluster from $PQ$ using $PQ.popLast()$. For
each cluster, the similarity index is recalculated using the
$SimilarityCalculator(c)$ function. If the recalculated similarity
index meets or exceeds the minimum similarity threshold, the
counter for high-similarity predictions is incremented. The
simulation stops when the counter reaches a predefined value
equal to $similarity\_range \cdot N$. This ensures that the
simulation is bounded by the specified similarity range. The
approximate recall, $recall\_approx$, is then computed as the ratio
of correctly predicted high-similarity clusters
($highSimilarityPrediction$) to the total number of actual
high-similarity clusters ($HighSimIndices$). Finally, the
$max\_size$ parameter, which determines the maximum number
of clusters to verify during the full verification phase, is
estimated using the formula:

\begin{equation} \mathrm{maxSize}= r_d \cdot \frac{1}{\mathrm{recallApprox}} \cdot \frac{\mathrm{HighSimIndices}}{N} \cdot \mathrm{totalClusters} \end{equation}

where $r_d$ is the desired recall rate. This formula adjusts the
maximum verification size based on the estimated recall, the
total number of high-similarity indices, and the desired recall
proportion.

Verification Phase (lines 59--78): The verification phase validates clusters by identifying those
with high similarity indices. The process begins by initializing an
empty priority queue $PQ$. For each cluster in the dataset, the
feature vector $u$ is computed using the predefined feature set
$F$. The trained machine learning model $M$ is then used to
calculate the probability $w_{s,t}$ of the cluster being a
high-similarity cluster. Each cluster, along with its probability,
is added to $PQ$, prioritizing clusters with higher probabilities.

The algorithm iteratively processes clusters in descending order
of their classification probabilities. The cluster $c$ with the
highest probability is removed from $PQ$ using $PQ.popLast()$,
and its actual similarity index $I_M$ is calculated. If $I_M$
exceeds or equals the minimum similarity threshold, the cluster
is added to the result set $L_R$. The process continues until the
counter reaches $max\_size$ or $PQ$ is empty. Once the priority
queue is exhausted or the size constraint is met, the result set
$L_R$, containing the verified high-similarity clusters, is
returned. This phase efficiently identifies the most promising
clusters while adhering to recall constraints and computational
limits.

\section{EXPERIMENTAL ANALYSIS}

The experimental analysis was conducted using a v28 TPU
machine with 334.56 GB of RAM, utilizing Python 3.9 as the
programming language. The experiments were implemented on
Google  Colab.  This  setup  allowed  to  evaluate  the  proposed
framework's  performance  on  diverse  datasets  with  varying
sizes and complexities effectively. Additionally, the algorithm
was executed without utilizing Triton for its computations, and
further  performance  comparisons  were  conducted
independently.

The experimental analysis evaluates the performance of the
proposed  framework  on  three  datasets  with  varying  sizes,
complexities,  and  average  cluster  sizes.  These  differences
allow us to explore the framework's scalability and its ability to
handle diverse conditions effectively.

Dataset  D1  includes  229,276  source  geometries  and
583,833 target geometries, forming a total of 295,481 clusters
with an average size of approximately six polygons per cluster.
Dataset D2 contains 210,483 source geometries and 2,898,899
target geometries, resulting in 654,196 clusters with an average
cluster  size  of  around  13  polygons.  Finally,  Dataset  D3
comprises  200,294  source  geometries  and  7,392,699  target
geometries, yielding 1,324,980 clusters with an average size of
34 polygons.

Regarding the distribution of similarity ranges, in Dataset
D1, most clusters fall within the 40-70\% similarity range, while
clusters  with  similarity  indices  above  90\% represent  a  very
small fraction---only 65 out of nearly 300,000 clusters. Dataset
D2 follows a similar pattern, with the majority concentrated in
the 50-70\% range and less than 0.5\% of clusters  in the 90-
100\%  similarity  range.  Dataset  D3  demonstrates  a  more
skewed distribution, with the majority in the 50-70\% range and
an even smaller proportion of high-similarity clusters.  These
results  suggest  that  highly  similar  clusters  are  rare,  which
underscores  the  importance  of  using  targeted  methods  to
identify them efficiently.

We also examined the dataset coverage required to identify
clusters within specific top percentages, such as the top 10\%,
30\%,  and  50\%.  In  Dataset  D1,  identifying  the  top  10\% of
clusters required processing $\sim$32\% of the dataset, while $\sim$45\%
and $\sim$60\% were processed to identify the top 30\% and 50\%,
respectively. Dataset D2 required $\sim$41\%, $\sim$55\%, and $\sim$70\%, and
Dataset  D3 required  $\sim$68\%,  $\sim$85\%,  and  $\sim$94\% for  the  same
ranges.  These  findings  demonstrate  the  adaptability  of  the
framework to prioritize computational resources, though larger
and more complex datasets naturally require greater processing
effort.

Further  insights  emerge  from  analyzing  the  analogy
between the percentage of clusters checked and the percentage
of clusters targeted.  For Dataset  D1, the ratio of checked to
targeted clusters was approximately 3.25 for the top 10\% and
decreased  as  the  target  percentage  increased,  reflecting
improved efficiency for broader ranges. Similarly, Dataset D2
exhibited a ratio of $\sim$4 for the top 10\%, decreasing to $\sim$1.8 for
the top 50\%. Dataset D3 began at $\sim$6.8 for the top 10\% but also
demonstrated decreasing ratios, reaching $\sim$1.9 for the top 50\%.
These  trends  indicate  that  while  larger  and  more  complex
datasets pose challenges,  the framework effectively allocates
effort based on the desired coverage.

Figure~4 presents six graphs that analyze the relationship
between  the  top  percentage  of  clusters  targeted  and  the
percentage of clusters checked or their ratios for three datasets
(D1, D2, and D3). Each row corresponds to a specific dataset,
showcasing the following:

\begin{itemize}
\item Top-Left (D1): The analogy of checked clusters to targeted
clusters for Dataset D1. It demonstrates how the percentage of
clusters  checked  scales  relative  to  the  targeted  clusters,
indicating  efficiency  in  narrowing  down  to  high-similarity
clusters.
\item Top-Middle (D1): The percentage  of  targeted  clusters  to
checked clusters for Dataset D1. This graph visualizes how the
effort required to identify specific top clusters scales with the
target percentage.
\item Top-Right  (D2):  The  analogy  of  checked  clusters  to
targeted clusters for Dataset D2. It highlights a similar trend for
Dataset D2, emphasizing the computational trade-offs required
to achieve targeted high-similarity clusters.
\item Bottom-Left  (D2):  The percentage of  targeted clusters  to
checked  clusters  for  Dataset  D2.  This  graph  compares  the
effort  to  achieve  targeted  clusters  against  the  percentage  of
clusters checked in Dataset D2.
\item Bottom-Middle (D3): The analogy of checked clusters  to
targeted clusters for Dataset D3. It shows how the ratio varies
for  Dataset  D3, which has larger  clusters,  demonstrating the
increasing computational cost with larger datasets.
\item Bottom-Right (D3): The percentage of targeted clusters to
checked  clusters  for  Dataset  D3.  It  outlines  how efficiently
clusters can be identified in Dataset D3, highlighting the effect
of average cluster size on computational requirements.
\end{itemize}

\begin{figure*}[t]
  \centering
  \includegraphics[width=0.95\textwidth]{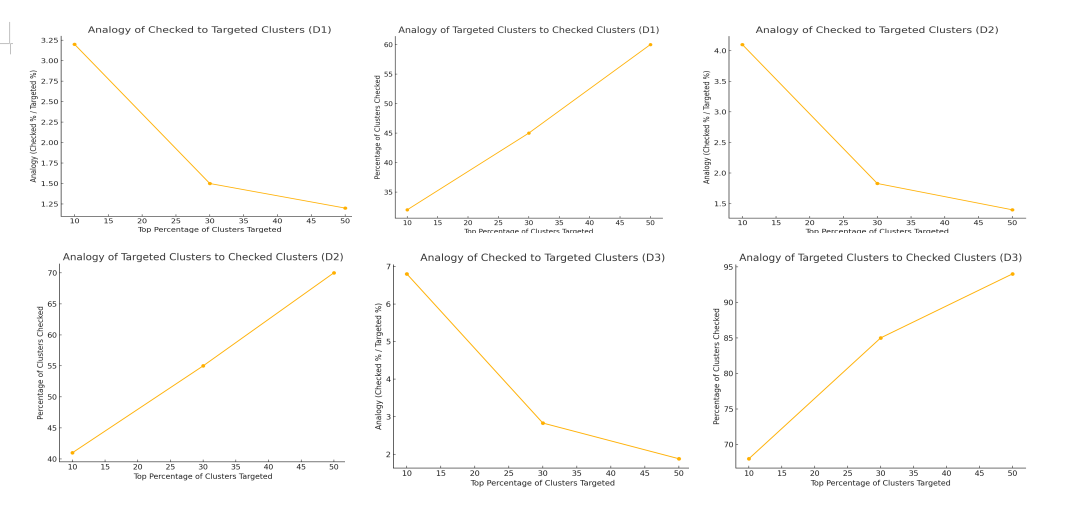}
  \caption{Analysis of the relation between the top percentage of clusters targeted and the percentage of clusters checked for datasets D1, D2 and D3.}
  \label{fig:wide_figure}
\end{figure*}

The relationship between average cluster size and dataset
coverage  reveals  a  direct  impact  on  computational
requirements.  Larger  clusters,  such  as  those  in  Dataset  D3,
demand  the  processing  of  a  much  larger  percentage  of  the
dataset to achieve the same recall levels as smaller clusters. For
example,  identifying  the top  50\% of  clusters  in  Dataset  D3
required processing $\sim$94\% of the dataset, compared to $\sim$60\% in
Dataset D1. This highlights the challenges of handling larger
cluster  sizes  and  underscores  the  necessity  of  employing
optimization techniques.

\subsection*{Insights on Shapely 2.0 and Triton}

To  further  support  these  analyses,  Shapely  2.0  offers
notable  advancements  in  geometric  operations.  By  enabling
vectorized  processing,  Shapely  2.0  processes  arrays  of
geometries  in  a  single  operation,  avoiding  the  overhead  of
handling each geometry sequentially.

\textbf{Performance  Improvements:} Vectorized  operations  in
Shapely  2.0 are  significantly faster,  offering  up  to  threefold
improvements in tasks such as calculating Minimum Bounding
Rectangles,  intersections,  and  unions.  These  enhancements
help reduce the computational time for critical pre-processing
steps in this study.

\textbf{Enhanced  GeoPandas  Integration:} The  improved
integration  with  GeoPandas  enables  faster  spatial  joins  and
overlays, particularly useful when handling large datasets like
those  analyzed  here.  This  compatibility  facilitates  efficient
preparation of geometries for similarity computations.

\textbf{Scalability for Large Datasets:} Shapely 2.0’s optimizations
make  it  feasible  to  process  large-scale  spatial  datasets
efficiently. While these improvements enhance computational
performance, they must be combined with targeted methods to
manage the overall computational workload effectively.

By integrating  these  tools  into the  proposed framework,  the
computational overhead associated with geometric operations
is reduced, making it feasible to address scalability challenges.
However,  while  these  advancements  are  promising,  their
application requires careful evaluation to ensure alignment with
the framework’s objectives.

In addition to the above, Triton was evaluated separately for
computing similarity metrics independently of the algorithm.
This evaluation involved directly computing similarity metrics
using  averages  to  gauge  potential  speedups.  Triton
demonstrated  speed  improvements  compared  to  standard
Python  code,  especially  when  handling  large-scale
parallelizable  operations.  While  these  results  are  promising,
they also highlight the need to reduce the number of required
computations  within  the  algorithm  to  ensure  that  Triton’s
parallel processing capabilities are utilized efficiently.

However, it is important to acknowledge that incorporating
Triton  into  the  broader  framework  requires  careful
optimization.  The  complexity  of  similarity  computations,
particularly in large datasets,  can still  pose challenges if  left
unoptimized. By leveraging Triton’s strengths in highly parallel
operations  while  minimizing  redundant  calculations,  future
work aims to fully harness its capabilities in conjunction with
tools like Shapely 2.0. These combined efforts hold promise for
achieving a scalable and efficient solution for identifying high-
similarity clusters in large datasets.


\bibliographystyle{IEEEtran}
\bibliography{references}

\end{document}